\documentclass[conference]{aiaa-pretty}
 \usepackage{float}
 \usepackage{caption}
 \usepackage{booktabs}
\DeclareCaptionType{equ}[][]
\usepackage{xcolor}

\usepackage{float}
\usepackage{aliascnt}
\newaliascnt{eqfloat}{equation}
\newfloat{eqfloat}{h}{eqflts}
\floatname{eqfloat}{}

\newcommand*{\ORGeqfloat}{}
\let\ORGeqfloat\eqfloat
\def\eqfloat{%
  \let\ORIGINALcaption\caption
  \def\caption{%
    \addtocounter{equation}{0}%
    \ORIGINALcaption
  }%
  \ORGeqfloat
}

\author[Wu, Li, and Chen]{ %
Pengcheng Wu\thanks{Joint PhD Student, Department of Aerospace Engineering, Member AIAA, \texttt{pwu@sdsu.edu,pcwupat@ucsd.edu}}\\
\textit{University of California San Diego, La Jolla, CA, 92093};\\
\textit{San Diego State University, San Diego, CA, 92182}\\
Lin Li\thanks{MS Student, Department of Biostatistics, \texttt{amandali@live.unc.edu}}\\ \textit{University of North Carolina at Chapel Hill, Chapel Hill, NC, 27599}\\
Junfei Xie \thanks{Assistant Professor, Department of Electrical and Computer Engineering, Member AIAA, \texttt{jxie4@sdsu.edu}},
Jun Chen\thanks{Assistant Professor, Department of Aerospace Engineering, Member AIAA, \texttt{jun.chen@sdsu.edu} (Corresponding Author)}\\
\textit{San Diego State University, San Diego, CA, 92182}}

\title{Probabilistic Guaranteed Path Planning for Safe Urban Air Mobility Using Chance Constrained $\text{RRT}^{*}$}

 \abstract{Safety is a critical concern for the success of urban air mobility, especially under dynamical and uncertain environment. This paper proposes a path planning algorithm based on $\text{RRT}^{*}$ in conjunction with chance constraints in the presence of uncertain obstacles. Chance constrained formulation for Gaussian distributed obstacles is developed by converting the probabilistic constraints to deterministic constraints in terms of distribution parameters. The probabilistic feasible region at every time step can be established through the simulation of the system state and the evaluation of convex constraints. Through establishing chance constrained $\text{RRT}^{*}$, the algorithm not only enjoys the benefits of sampling-based algorithms but also incorporates uncertainty into the formulation. Simulation results demonstrate that the planning for a trajectory connecting the starting and goal point in accordance with the requirement of probabilistic obstacle avoidance can be achieved by the utilization of this algorithm.}

\begin{document}
\maketitle
\section{Introduction}
Urban air mobility, a safe and efficient air traffic transportation system around metropolitan areas, is now increasingly drawing attention \cite{thipphavong2018urban}. One concern is that the high density air traffic system may come across certain disturbance like environmental winds, affecting the operation of the aircraft system. Also, for the safety of a multi-agent system formed by multiple aircrafts, the possibility of collision between each other needs to be considered. For a certain aircraft in the system, such environmental disturbance and all the other aircrafts in the system can be viewed as uncertain obstacles. Therefore, it is important to develop an algorithm which enables identifying feasible paths for the aircraft system in the presence of uncertain obstacles. To reach safe and reliable path planning in reality where uncertain obstacles may be present, we expect that the information of the uncertainty be included into the planning problem.\\

A way to model uncertain obstacles while planning trajectories is through chance constraints. The chance constraint formulation proposed by Blackmore et al. assumes that all the obstacles have known, static locations\cite{blackmore2008robust,blackmore2006probabilistic}. Some extensions based on Blackmore’s work have been further explored. Toit extended the chance constrained optimization framework to consider other kinds of uncertainty, such as collision avoidance between uncertain agents\cite{du2010robot}. Luders et al. incorporated within the formulation the uncertainty combining system state distribution and location of dynamic obstacles\cite{luders2010chance}, where those dynamic obstacles they referred to are only assumed to follow known, deterministic paths.\\
 
A rapidly-exploring random tree (RRT), an incremental sampling-based algorithm first proposed by LaValle, is intended for path planning problems that involve obstacles and differential constraints\cite{lavalle2006planning,lavalle1998rapidly}. Compared with traditional motion planning algorithms, like potential field method, genetic algorithm, the neural network method, etc., the RRT algorithm is able to implement collision checking for widespread sampling points in the state space without space modeling. RRT algorithm can be regarded as a component incorporated into the development of a variety of different planning algorithms. Some improvements have also been made based on the variations of the original RRT algorithm\cite{frazzoli2002real,kuwata2009real}. Karaman et al. developed $\text{RRT}^{*}$ algorithm, guaranteeing the asymptotic optimality while maintaining a tree structure like original RRT algorithm \cite{karaman2011sampling}.\\

In this paper, the chance constrained formulation allowing for probability distributed obstacles using $\text{RRT}^{*}$ algorithm is investigated. The remaining part of the article is organized as follows. In section 2, the statement of the problem explored in this paper is established. In section 3, the chance constraint formulation which allows for probability distributed obstacles as uncertain constraints is fully discussed. In section 4, the chance constraint formulation is incorporated into the $\text{RRT}^{*}$ algorithm, and in turn the chance constrained $\text{RRT}^{*}$ algorithm is presented. In section 5, the feasibility of the proposed algorithm is verified through numerical simulation. Finally, we conclude this article in section 6.\\

\section{Problem Statement}
Consider a linear time invariant system at the discrete-time domain, which is modeled as
\begin{equation}
\begin{array}{l}{\boldsymbol{x}_{t+1}=\boldsymbol{A} \boldsymbol{x}_{t}+\boldsymbol{B} \boldsymbol{u}_{t}} \\ {\boldsymbol{y}_{t}=\boldsymbol{C} \boldsymbol{x}_{t}}\end{array}
\end{equation}
where $\boldsymbol{x}_{t} \in R^{n_{x}}$ represents the system state vector, $\boldsymbol{u}_{t} \in R^{n_{u}}$  represents the system input vector, and $\boldsymbol{y}_{t}$ represents the system output vector.\\

There may be constraints acting on the system state and input control sequence, which are assumed to take the following form

\begin{equation}
\begin{array}{l}{\boldsymbol{x}_{t} \in \neg \mathrm{X}_{t}:= \neg \left(\bigcup_{i=1}^{B} \mathrm{X}_{t i}\right)} \\ {\boldsymbol{u}_{t} \in \mathrm{U}_{t}}\end{array}
\end{equation}

where the set $\mathrm{X}_{t i} \subseteq R^{n_{x}}(i=1, \ldots, B)$ denotes the risk domain under Gaussian distribution at a given risk level, associated with B probability distributed obstacles which require avoidance. The time dependence of $\mathrm{X}_{t}$  allows both static and dynamic obstacles included. In addition, $\mathrm{U}_{t} \subseteq R^{n_{u}}$  denotes the constraints imposed on the input control sequence.\\

Each obstacle is assumed to be distributed as a two dimensional Gaussian distribution. In other words, the location of each obstacle is represented through a random vector $\boldsymbol{X}=(x, y)^{\mathrm{T}}$ , whose probability distribution is Gaussian with certain mean and covariance values. The probability density function (PDF) of $\boldsymbol{X}$ is defined as, 

\begin{equation}
f(\boldsymbol{X})=\frac{1}{(2 \pi)^{d / 2}|\boldsymbol{\Sigma}|^{1 / 2}} \exp \left[-\frac{1}{2}(\boldsymbol{X}-\boldsymbol{\mu})^{\mathrm{T}} \boldsymbol{\Sigma}^{-1}(\boldsymbol{X}-\boldsymbol{\mu})\right]
\end{equation}

where\\
$\boldsymbol{X}=(x, y)^{\mathrm{T}}$, a random vector with two random variables as components\\
$d$, a scalar which denotes the dimension of the random vector. In this case, $d=2$ \\
$\boldsymbol{\mu}=(\mu_{1}, \mu_{2})^{\mathrm{T}}$, a vector which denotes the mean values of two random variables respectively\\
$\boldsymbol{\Sigma}=\left(\begin{array}{ll}{\sigma_{11}} & {\sigma_{12}} \\ {\sigma_{21}} & {\sigma_{22}}\end{array}\right)$, a symmetric positive definite matrix termed the covariance matrix. Note that the elements $\sigma_{11}$ and $\sigma_{22}$  represent the variances of two variables respectively, while $\sigma_{12}$ or $\sigma_{21}$  represents the correlation coefficients between those two variables. Moreover, if $\sigma_{12}$ \textgreater 0, the random variables $x$ and $y$ are positively correlated; if  $\sigma_{12}$ \textless 0, they are negatively correlated; otherwise, they are definitely not correlated.\\

Given that each obstacle is probability distributed, a risk level $\alpha$  and a corresponding risk domain can be introduced to quantify the probability for the vehicle to avoid the collision with uncertain obstacles. Basically, the risk domain of Gaussian distribution at a given risk level is a circle or an ellipse. As a result,  $\mathrm{X}_{t i} \subseteq R^{n_{x}}(i=1, \ldots, B)$  represents a convex set with known shape and boundary. It will be explored in further detail when it comes to the next section. \\

The primary objective of the motion planning problem presented in this paper is to find a feasible path for a vehicle starting from the initial state to the goal domain $D_{goal} \subseteq R^{n_{x}}$  within minimum time. That is, to realize
\begin{equation}
\begin{array}{l}{\lim _{t \rightarrow t_{\text {gal}}} \boldsymbol{x}_{t} \in D_{\text {goal}}} \\ {t_{\text {goal}}=\inf \left\{t: \boldsymbol{x}_{t} \in D_{\text {goal}}\right\}}\end{array}
\end{equation}
while guaranteeing all the constraints and obstacle avoidance requirements are satisfied at each time step. A penalty function $\sum_{t=0}^{t_{goal}} \phi\left(\boldsymbol{x}_{t}, X_{t}, U_{t}\right)$  can also be introduced to avoid some undesirable circumstances, like radical proximity to the boundaries of constraints. With this, the motion planning problem introduced above can be formulated. \\

Given the initial state $\boldsymbol{x}_{0} \in R^{n_{x}}$  and constraints, figure out the input control sequence $\boldsymbol{u}_{t} \in R^{n_{u}}$  such that

\begin{equation}
\left\{\begin{array}{l}{\min J=t_{g o a l}+\sum_{t=0}^{t_{m a t}} \phi\left(\boldsymbol{x}_{t}, X_{t}, U_{t}\right)} \\ {\text { s.t. }} \\ {\boldsymbol{x}_{t+1}=\boldsymbol{A} \boldsymbol{x}_{t}+\boldsymbol{B} \boldsymbol{u}_{t}} \\ {\boldsymbol{x}_{t_{0}}=\boldsymbol{x}_{0}} \\ {\boldsymbol{x}_{t} \in \neg X_{t}} \\ {\boldsymbol{u}_{t} \in U_{t}} \\ {\forall t \in\left\{t_{0}, \ldots, t_{g o a l}\right\}}\end{array}\right.
\end{equation}\\

\section{Chance Constraints for Gaussian Distributed Obstacles} \label{ssec:criteria}
In this section, we take into consideration how probability distributed obstacles can be settled, as an extension of the traditional chance constraint formulation. In doing so, the uncertainty of obstacles found in common path planning scenarios is able to be incorporated within the chance constraint formulation.\\

Suppose that our objective is to ensure that the probability of collision with any obstacles for a given time step is less than a threshold value $\Delta$. To make the problem tractable for path planning algorithms, it is important to transform the probabilistic requirement into deterministic constraints.\\

Let $\boldsymbol{X}$ be a random vector with the size of $d$ by 1, which obeys a $d$-dimensional Gaussian distribution $N_{d}(\boldsymbol{\mu}, \boldsymbol{\Sigma})$. As $\boldsymbol{X} \sim N_{d}(\boldsymbol{\mu}, \boldsymbol{\Sigma})$ , then we have $(\boldsymbol{X}-\boldsymbol{\mu}) \sim N_{d}(\boldsymbol{0}, \boldsymbol{\Sigma})$. Thus,  $T^{2}=(\boldsymbol{X}-\boldsymbol{\mu})^{\mathrm{T}} \boldsymbol{\Sigma}^{-1}(\boldsymbol{X}-\boldsymbol{\mu}) \sim \chi^{2}_{d}$, where $T^{2}$ is a random scalar and $\chi^{2}_{d}$ stands for a chi-squared distribution with $d$ degrees of freedom \cite{snijders2011multilevel}.\\

The cumulative distribution function (CDF) of $\chi^{2}_{d}$ is 

\begin{equation}
F(x, d)=\frac{\gamma(\frac{d}{2},\frac{x}{2})}{\Gamma(\frac{d}{2})}=P\left(\frac{d}{2}, \frac{x}{2}\right)
\end{equation}

where $P\left(\frac{d}{2}, \frac{x}{2}\right)$ is the regularized gamma function. Especially, if $d=2$, this CDF has a simple form

\begin{equation}
F(x, 2)=1-e^{-\frac{x}{2}}
\end{equation}

Because $F(x, d): R^{+} \rightarrow [0, 1)$ is a bijective mapping, we can find the inverse mapping of $F(x, d)$, which is defined as $F^{-1}(x, d): [0, 1) \rightarrow R^{+}$.\\

Given a risk level $\alpha$ ,we can find a critical value $F^{-1}(1-\alpha, d)$ such that $\text{Pr}\left(T^{2}>F^{-1}(1-\alpha, d)\right)=\alpha$. Therefore, $T^{2}=({\boldsymbol{X}}-\boldsymbol{\mu})^{\mathrm{T}} \boldsymbol{\Sigma}^{-1}({\boldsymbol{X}}-\boldsymbol{\mu})=F^{-1}(1-\alpha, d)$ forms the boundary of the risk domain $\mathrm{X}_{t i} \subseteq R^{n_{x}}(i=1, \ldots, B)$, which is derived from the Gaussian distribution of $\boldsymbol{X}$ at the risk level $\alpha$.\\

In fact, the expression $({\boldsymbol{X}}-\boldsymbol{\mu})^{\mathrm{T}} \boldsymbol{\Sigma}^{-1}({\boldsymbol{X}}-\boldsymbol{\mu})$  is a typical quadratic form in linear algebra. Particularly, if $d=2$ and thereby $\boldsymbol{X} \sim N_{2}(\boldsymbol{\mu}, \boldsymbol{\Sigma})$, then $T^{2}=({\boldsymbol{X}}-\boldsymbol{\mu})^{\mathrm{T}} \boldsymbol{\Sigma}^{-1}({\boldsymbol{X}}-\boldsymbol{\mu})=F^{-1}(1-\alpha, 2)$ can be geometrically interpreted as a circle or an ellipse, dependent on the given values of the mean vector $\boldsymbol{\mu}$ and the covariance matrix $\boldsymbol{\Sigma}$ . That is to say, the risk domain $\mathrm{X}_{t i} \subseteq R^{n_{x}}(i=1, \ldots, B)$  represents a convex set with known shape and boundary.\\

The B Gaussian distributed obstacles are represented through the quadratic inequalities, 

\begin{equation}
\vee\left(\left(\boldsymbol{x}_{t}-\boldsymbol{\mu}_{i}\right)^{\mathrm{T}} \boldsymbol{\Sigma}_{i}^{-1}\left(\boldsymbol{x}_{t}-\boldsymbol{\mu}_{i}\right) \leq F^{-1}(1-\alpha, d)\right) \quad i=1, \ldots, B
\label{disjunction}
\end{equation}

To avoid all the obstacles at each time step, the system must satisfy B quadratic inequalities at the same time.
\begin{equation}
\wedge\left(\left(\boldsymbol{x}_{t}-\boldsymbol{\mu}_{i}\right)^{\mathrm{T}} \boldsymbol{\Sigma}_{i}^{-1}\left(\boldsymbol{x}_{t}-\boldsymbol{\mu}_{i}\right) > F^{-1}(1-\alpha, d)\right) \quad i=1, \ldots, B,   \quad t \in\left\{0, \ldots, t_{g o a l}\right\}
\label{conjunction}
\end{equation}                                                                                                                                               
Consider the problem of avoiding the $j_{th}$ obstacle on the $t_{th}$ time step. To avoid the obstacles with a risk probability less than $\alpha$ , it is necessary to not satisfy any one of the quadratic inequalities given in Eq (\ref{disjunction}). Conversely, any violation of the quadratic inequalities inequalities given in Eq (\ref{conjunction}) will lead to the failure of collision avoidance. In fact, it is the case that,
	  
\begin{equation}
\begin{array}{l}{\text{Pr}(\text { collision })=\text{Pr}\left(\vee\left(\left(\boldsymbol{x}_{t}-\boldsymbol{\mu}_{i}\right)^{\mathrm{T}} \boldsymbol{\Sigma}_{i}^{-1}\left(\boldsymbol{x}_{t}-\boldsymbol{\mu}_{i}\right) \leq F^{-1}(1-\alpha, d)\right)\right)}\\ 

{=\sum_{i=1}^{B} \text{Pr}\left(\left(\boldsymbol{x}_{t}-\boldsymbol{\mu}_{i}\right)^{\mathrm{T}} \boldsymbol{\Sigma}_{i}^{-1}\left(\boldsymbol{x}_{t}-\boldsymbol{\mu}_{i}\right) \leq F^{-1}(1-\alpha, d)\right)-\text{Pr}\left(\wedge\left(\left(\boldsymbol{x}_{t}-\boldsymbol{\mu}_{i}\right)^{\mathrm{T}} \boldsymbol{\Sigma}_{i}^{-1}\left(\boldsymbol{x}_{t}-\boldsymbol{\mu}_{i}\right) \leq F^{-1}(1-\alpha, d)\right)\right)} \\ 

{\leq\sum_{i=1}^{B} \text{Pr}\left(\left(\boldsymbol{x}_{t}-\boldsymbol{\mu}_{i}\right)^{\mathrm{T}} \boldsymbol{\Sigma}_{i}^{-1}\left(\boldsymbol{x}_{t}-\boldsymbol{\mu}_{i}\right) \leq F^{-1}(1-\alpha, d)\right)} \\ {i=1, \ldots, B}\end{array}
\end{equation}
To ensure that the probability of collision is less than $\alpha$, it is only required to show that the significance level  of the Gaussian distribution is satisfied with  . 
 \begin{equation}
\begin{array}{l}{\text{Pr}(\text { collision }) \leq \sum_{i=1}^{B} \text{Pr}\left(\left(\boldsymbol{x}_{t}-\boldsymbol{\mu}_{i}\right)^{\mathrm{T}} \boldsymbol{\Sigma}_{i}^{-1}\left(\boldsymbol{x}_{t}-\boldsymbol{\mu}_{i}\right) \leq F^{-1}(1-\alpha, d)\right) \leq B \alpha \leq \Delta} \\ {i=1, \ldots, B}\end{array}
\end{equation}

According to the discussion above, the probabilistic collision avoidance can be satisfied on condition that the planned paths of the vehicles system don't intersect with the risk domain of the B obstacles at the risk level $\alpha$ satisfying probability inequality $B \alpha \leq \Delta$.\\

\section{Chance Constrained $\text{RRT}^{*}$ Algorithm}
$\text{RRT}^{*}$ algorithm is a sampling-based motion planning algorithm intended for the efficient search for high dimensional space. Karaman et al. proved that for sampling-based motion planning algorithms, as the sampling points of RRT algorithm tend to infinity, the probability of its convergence to the optimal solution becomes zero\cite{karaman2011sampling}. Alternatively, they proposed $\text{RRT}^{*}$ algorithm which ensures the asymptotic optimality on the basis of the original RRT algorithm. This algorithm improves the way of parent node selection, and introduces the notion of cost function to select the node as the parent node which has the minimum cost within the neighborhood of the node extended. At the same time, nodes in the existing tree will be reconnected after each iteration, so as to ensure the computational complexity and asymptotic optimality.\\

$\text{RRT}^{*}$ algorithm is suited for path planning problems concerning obstacles and differential constraints. It is able to plan a trajectory connecting the starting and goal point for multi-agent systems in the presence of obstacles and constraints. Based on that, the chance constrained $\text{RRT}^{*}$ algorithm is considered as an extension of standard $\text{RRT}^{*}$ algorithm in the sense that it applies trajectory-wise constraints checking, allowing for the incorporation of probabilistic constraints. While standard $\text{RRT}^{*}$ algorithm grows a tree of states as are known to be feasible, the chance constrained $\text{RRT}^{*}$ algorithm creates a tree of states satisfying constraints subject to uncertain obstacles with a certain probability. \\

The chance constrained $\text{RRT}^{*}$ tree expansion step, used to incrementally grow the tree, is given in Algorithm 1. It utilizes the chance constraint formulation developed in the last section, which applies a fixed probability bound $\Delta/B$ across all obstacles, lead to deterministic constraints able to be computed at each time step. It works for checking probabilistic feasibility of planned trajectories generated by the chance constrained $\text{RRT}^{*}$ algorithm. With the help of chance constraint formulation, one can figure out whether the deterministic constraints are satisfied through judging the relative positions between the simulated trajectories and the risk domain of the obstacles for a given mean $\boldsymbol{\mu}$  and  covariance $\boldsymbol{\Sigma}$ at a risk level $\alpha$. The simulated trajectory is adopted only if these chance constraints are satisfied.\\

\begin{table}[H] 
 \caption*{}
 \begin{tabular}{l} 
  \toprule 
   \textbf{Algorithm 1: Chance Constrained RRT*}\\ 
  \midrule 
  1. $ V` \leftarrow V; E` \leftarrow E;$  \\ 
   2. $q_{nearest} \leftarrow Nearest(G,q);$\\ 
   3. $q_{new} \leftarrow Steer(q_{nearest},q); $ \\ 
     4. $\textbf{if }\; ObstacleFree(q_{nearest},q_{new}) \textbf{ then}$  \\ 
      5. $\quad \quad \quad V` \leftarrow V` \bigcup \{ q_{new}\} ;$   \\
      6.$\quad \quad \quad q_{min}\leftarrow q_{nearest}$\\
      7.$\quad \quad \quad C_{near}\leftarrow Near(G, q_{new}, |V|);$\\
      8.$\quad \quad \quad \textbf{for}\; all\; q_{near}\in C_{near} \textbf{do}$\\
      9.$\quad \quad \quad \quad \textbf{if }\; ObstacleFree(q_{near},q_{new})\textbf{ then}$\\
       10.$\quad \quad \quad \quad \quad c`\leftarrow Cost(q_{near})+pc(Line(q_{near},q_{new}))$\\
       11. $\quad \quad \quad \quad \quad \textbf{ if}\; c`<Cost(q_{new})\; \textbf{ then}$\\
       12. $\quad \quad \quad \quad \quad \quad q_{min}\leftarrow q_{near}$\\
       13. $\quad \quad \quad E` \leftarrow E` \bigcup \{ (q_{nearest},q_{new})\}$  \\
       14. $\quad \quad \quad \textbf{for}\; all\; q_{near}\in C_{near} - \{ q_{min}\}  \textbf{do}$\\
       15.$\quad \quad \quad \quad \textbf{if }\; ObstacleFree(q_{new},q_{near}) \wedge Cost(q_{near}) > Cost(q_{new})+c(Line(q_{new},q_{near})) \textbf{ then}$    \\
       16.  $\quad \quad \quad \quad \quad q_{parent}\leftarrow Parent(q_{near})$    \\
       17. $\quad \quad \quad \quad \quad E` \leftarrow E` - \{(q_{parent},q_{near})\}; E` \bigcup E' - \{(q_{new},q_{near})\};$  \\
       18. $ \textbf{Return}\; G`=(V`,E`) $\\
  \bottomrule 
 \end{tabular} 

\end{table}

\begin{table}
 \caption{table}
\begin{tabular}{l}

123

\end{tabular}
\end{table}
For the flight safety of a multi-agent system, in addition to the environmental obstacles, the possibility of collision between any two vehicles also needs to be considered. At the beginning of every new time step, when we are ready to extend the trajectory for a certain vehicle in the multi-agent system, the possible trajectories for any other vehicles within the same time step may haven't been determined yet. In view of this, for the particular vehicle being prepared to be planned, the positions of other vehicles within current time step are uncertain. Thus, for any vehicle in the system, all the other vehicles can be modeled as probability distributed obstacles. In doing so, the problem of agents uncertainty can be transformed to a special scenario under the framework of chance constrained formulation presented above. With this new form of obstacles being incorporated into the planning problem, the chance constrained $\text{RRT}^{*}$ algorithm proposed above can also apply to this scenario. Related work on the chance constrained $\text{RRT}^{*}$ algorithm for a multi-agent system will be fully discussed in the final version of this article in the future.\\

\section{Simulation Results}
First, consider a one-agent system in the presence of Gaussian distributed obstacles. Simulation results are presented, demonstrating the chance constrained $\text{RRT}^{*}$ algorithm   is effective in generating feasible trajectories for motion planning problems together with satisfying probabilistic constraints. \\

\begin{figure}[H]
    \centering
    \includegraphics[width=3in]{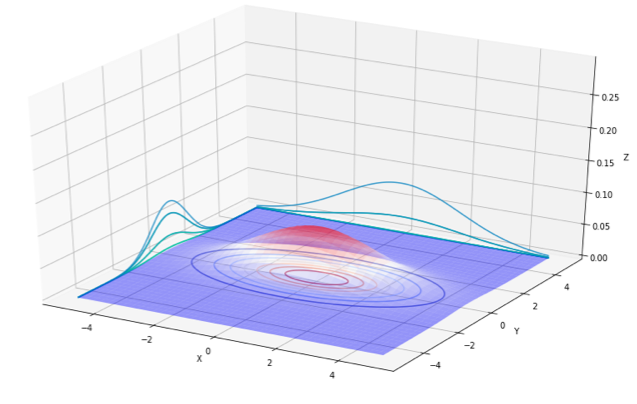}
   
    \label{fig:my_label}
    \caption{The probability density function of $\boldsymbol{X}=(x, y)^{\mathrm{T}}$ }
    \label{fig1}
\end{figure}

The probability density function of $\boldsymbol{X}=(x, y)^{\mathrm{T}}$ which obeys a standard two dimensional Gaussian Distribution is shown in Figure \ref{fig1}. For simplicity, suppose that each of four obstacles obeys a two dimensional Gaussian distribution with different means $\boldsymbol{\mu}_{1}=(1, 1)^{\mathrm{T}},  \boldsymbol{\mu}_{2}=(1, 5)^{\mathrm{T}},  \boldsymbol{\mu}_{3}=(6, 1)^{\mathrm{T}},  \boldsymbol{\mu}_{4}=(6, 8.8)^{\mathrm{T}}$  and a same covariance matrix $\boldsymbol{\Sigma}=\left({\frac{2}{3}}, \; {0}\;;\; {0}, \; {\frac{1}{6}}\right)$. The starting point and goal point are $(0, 0),  (6, 10)$ respectively.

\begin{figure}[H]
 \begin{center}
\includegraphics[width=3in]{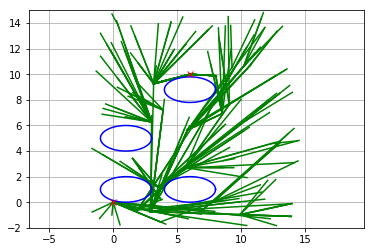}
 \end{center}
 \caption{The tree generated by chance constrained $\text{RRT}^{*}$ algorithm}
 \label{fig2}
\end{figure}

\begin{figure}[H]
 \begin{center}
  \includegraphics[width=3in]{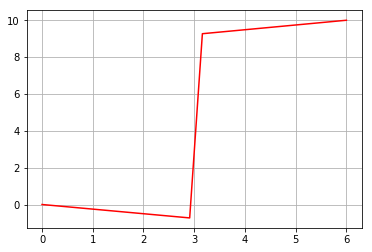}
 \end{center}
 \caption{The trajectory planned by chance constrained $\text{RRT}^{*}$ algorithm}
 \label{fig3}
\end{figure}

Consider performing the chance constrained $\text{RRT}^{*}$ algorithm on this scenario with a risk level $\alpha=5\%$. Figure \ref{fig2}-\ref{fig3} show the tree generated by this algorithm after tree expansion. It turns out that the trajectory connecting the starting and goal point can be planned by this algorithm, with probability distributed obstacles being successfully avoided. The risk domain at the risk level $\alpha$, which is an ellipse, is shown for every obstacle in Figure \ref{fig2}. This scenario demonstrates the ability of the chance constrained $\text{RRT}^{*}$ algorithm to incorporate probabilistic obstacles into its computation. \\

\begin{figure}[H]
 \begin{center}
\includegraphics[width=3in]{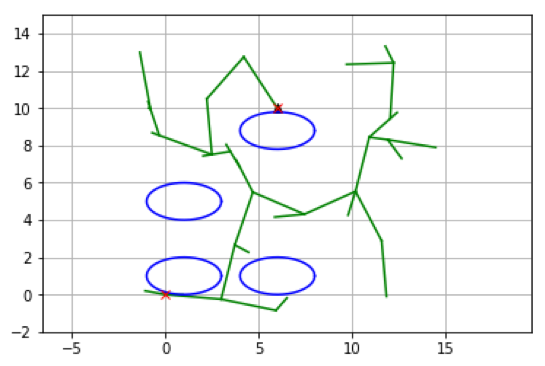}
 \end{center}
 \caption{The tree generated by chance constrained RRT algorithm}
 \label{fig4}
\end{figure}

\begin{figure}[H]
 \begin{center}
  \includegraphics[width=3in]{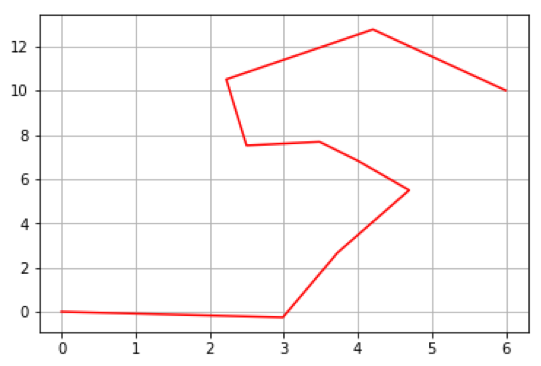}
 \end{center}
 \caption{The trajectory planned by chance constrained RRT algorithm}
 \label{fig5}
\end{figure}

By comparison, the chance constrained RRT algorithm can also be applied to find a feasible trajectory. The tree generated by chance constrained RRT algorithm is shown in Figure \ref{fig4}-\ref{fig5}. Compared with RRT, the chance constrained $\text{RRT}^{*}$ algorithm can guarantee the asymptotic optimality while maintaining a tree structure. 

\section{Conclusions}                   
In this paper, the chance constrained $\text{RRT}^{*}$ algorithm, a sampling-based motion planning method, is presented and fully discussed, allowing for the incorporation of the uncertainty attributed to probability distributed obstacles. The chance constraints are formulated through representing the probabilistic constraints as deterministic ones. The risk domain of a random vector which obeys a Gaussian distribution can be obtained through transforming it to the problem involving chi-square distribution, with which one can ensure those constraints are satisfied through judging the relative positions between simulated trajectories and the risk domain. As an extension, the framework of chance constrained $\text{RRT}^{*}$ algorithm can also apply to a multi-agent system provided that uncertain agents are modeled as dynamical obstacles. As demonstrated through simulation results, the path planning for a trajectory connecting the starting and goal point in accordance with the requirement of probabilistic collision avoidance can be achieved by using this algorithm.


\medskip
\nocite{*}
\bibliographystyle{aiaa}
\bibliography{refs}

\end{document}